%% file: main.tex
\documentclass[11pt]{article}

\usepackage[final]{acl}

\usepackage{times}
\usepackage{latexsym}

\usepackage[T1]{fontenc}

\usepackage[utf8]{inputenc}

\usepackage{microtype}

\usepackage{inconsolata}

\usepackage{graphicx}

\input{macros}
\input{packages}

\title{Meronymic Ontology Extraction via Large Language Models}

\author{
 \textbf{Dekai Zhang\textsuperscript{1}},
 \textbf{Simone Conia\textsuperscript{2}} \and
 \textbf{Antonio Rago\textsuperscript{1,3}}
\\
 \textsuperscript{1} Imperial College London, UK \\
 \textsuperscript{2} Sapienza University of Rome, Italy \\
 \textsuperscript{3} King's College London, UK
\\
 \small{
   \textbf{Correspondence:} \href{mailto:antonio.rago@kcl.ac.uk}{antonio.rago@kcl.ac.uk}
 }
}

\begin{document}
\maketitle

\begin{abstract}

    Ontologies have become essential in today’s digital age as a way of organising the vast amount of readily available unstructured text. 
    In providing formal structure to this information, ontologies have immense value and application across various domains, e.g., e-commerce, where countless product listings 
    necessitate proper product organisation. 
    However, the manual construction of these ontologies is a time-consuming, expensive and laborious process.
    In this paper, we harness the recent advancements in large language models (LLMs) to develop a fully-automated method of extracting product ontologies, in the form of meronomies, from raw review texts. 
    We demonstrate that the ontologies produced by our method surpass an existing, BERT-based baseline when evaluating using an LLM-as-a-judge. 
    Our investigation provides the groundwork for LLMs to be used more generally in (product or otherwise) ontology extraction.

\end{abstract}


\section{Introduction}
\label{sec:introduction}

The proliferation of the internet has led to an ever-increasing amount of unstructured text data, presenting both opportunities and challenges in harnessing this wealth of information. 
Ontologies are one method of organising the information from unstructured text and formally representing it as a graph. 
Such data representations are important in numerous downstream tasks, such as review aggregation \cite{Konjengbam_18}, sentiment analysis \cite{Schouten_18} and product question answering \cite{Kulkarni_19}.
Often, these ontologies can be as simple as consisting only of \emph{part-whole} relations, i.e. they are \emph{meronomies}, which have been shown to be particularly useful in review aggregation contexts concerning product \cite{Oksanen_21}, travel \cite{Rago_25} or movie \cite{Cocarascu_19} recommendation.
However, the manual construction of (even simple meronymic) ontologies is not only time-consuming but often demands domain expertise.

To this end, research efforts have been directed at learning these ontologies from data, notably via deep learning methods (see \citet{Al-Aswadi_20} for an overview). 
More recently, large language models (LLMs) \cite{Min_24}, which have revolutionised the field of NLP, have also found application in this domain with its increasing adoption in key ontology learning tasks of term and relation extraction \cite{Ye_22}. 
However, most of these efforts have been concentrated on extracting taxonomic \AR{(\emph{is})}, rather than meronymic \AR{(\emph{is part of})}, relations. 
\AR{While taxonomies}
\DZ{are routinely applied in increasingly specific classification of entities, in review aggregation contexts, meronomies are particularly useful for representing features of entities.}
\AR{For example,} 
\citet{Oksanen_21} 
introduce a method for extracting meronomies using two fine-tuned BERTs \cite{Devlin_19}. However, the method still relies on some manual annotation from humans.
Further, the effective evaluation of these meronymic ontologies is also non-trivial, since no ground-truth examples exist for this task currently.

\begin{figure*}[t]
    \centering
    \includegraphics[width=1\textwidth]{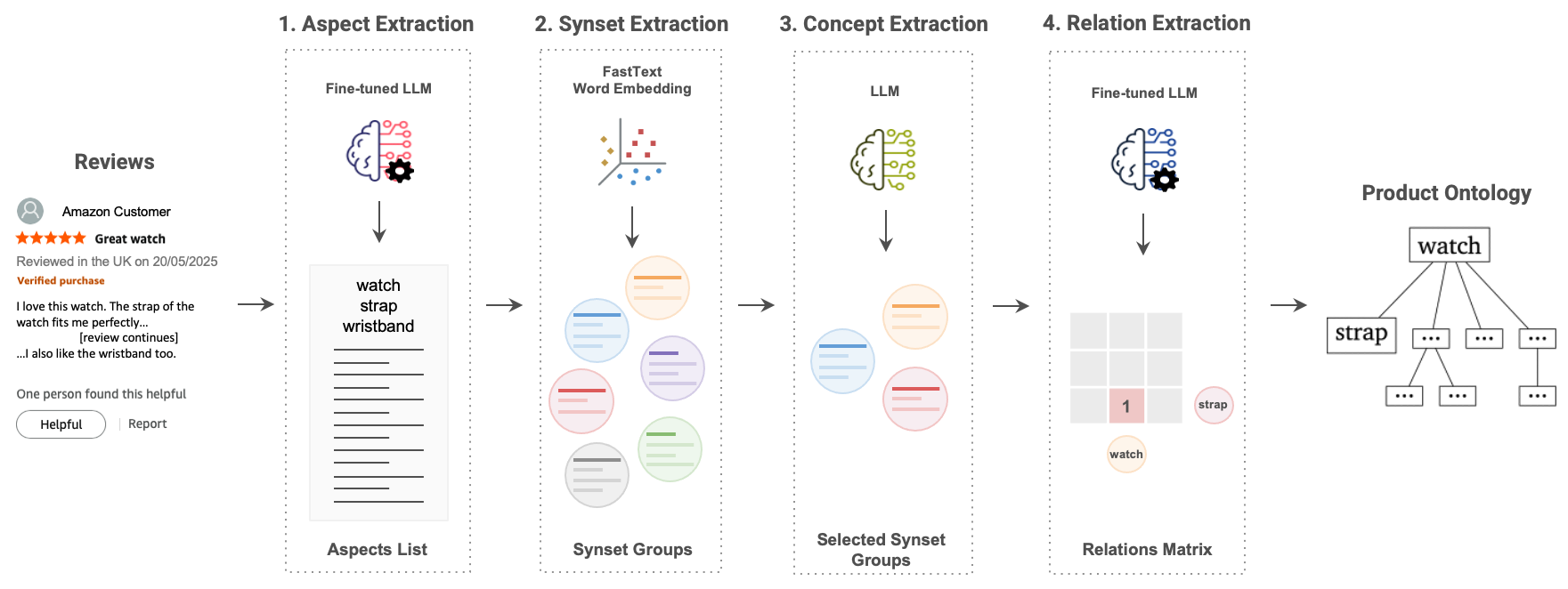}
    \caption{Our complete pipeline for extracting meronymic ontologies from product reviews using LLMs, consisting of the four tasks of aspect, synset, concept and relation extraction. \AR{In the example shown here, we begin with a review from a customer. The first task comprises the extraction of the relevant aspects from the review, e.g. \emph{watch}, \emph{strap} and \emph{wristband}, with a fine-tuned LLM. The second task then uses a word embedding to extract synset groups, e.g. discarding \emph{wristband} since it is a synonym of \emph{strap}.} \DZ{The third task is to select the most relevant of these with an LLM to obtain the concepts.} \AR{Finally, in the last task, a fine-tuned LLM extracts the relations between concepts, i.e. identifying that \emph{strap} is part of (in meronymic relation with) \emph{watch}, resulting in the final product ontology.}}
    \label{fig:pipeline}
\end{figure*}

In this paper, we harness the recent advancements in LLMs to make a number of contributions improving on the pipeline for ontology extraction proposed by \citet{Oksanen_21}. Concretely:

    %
    \noindent $\bullet$ We introduce a fully-automated method which uses LLMs for the extraction of meronymic ontologies, which generalises across different product categories (see Figure~\ref{fig:pipeline}). \\
    %
    \noindent $\bullet$ We propose a novel method for the empirical evaluation of the individual tasks 
    in meronymic ontology extraction using LLM-as-a-judge. \\
    %
    \noindent $\bullet$ We \DZ{empirically evaluate our LLM-based approach against a BERT-based method \citep{Oksanen_21}, finding significant gains in relevance.}\footnote{Source code: \url{https://github.com/dkaizhang/llm-meronomy}}
    %





\section{Meronymic Ontology Extraction via LLMs}
\label{sec:main}

In this section, we first describe the dataset used in both the method and the evaluation
, as well as the LLMs used for the different extraction tasks
.
We then outline our approach to the four tasks \AR{constituting} the meronymic ontology extraction pipeline, namely: 
aspect extraction
,
synset extraction
,
concept extraction 
and
relation extraction
.
Our approach 
is 
\AR{illustrated and motivated} in Figure \ref{fig:pipeline}.

\paragraph{Dataset}

The product reviews were obtained 
from the Amazon Reviews 2023 dataset \cite{Hou_24}. For the generation of the product ontologies, we selected the same five products as \citet{Oksanen_21} to allow for \AR{a fair} comparison
. These products were chosen to represent the diverse selection of products 
on Amazon: 
video games
, 
televisions
, 
necklaces/watches 
and stand mixers
. We randomly selected 100,000 reviews for each product, except for stand mixers, where only 26,464 reviews were available. 
The full dataset of 100,000 reviews was used for the synset extraction task. However, for the aspect and relation extraction tasks (which relied on LLMs), we limited the input to only 1,000 reviews due to the processing time of LLMs. 

\paragraph{Deployed LLMs}

For all our extraction tasks, we used the Mistral-7B-Instruct-v0.2 LLM\footnote{\href{https://huggingface.co/mistralai/Mistral-7B-Instruct-v0.2}{https://huggingface.co/mistralai/Mistral-7B-Instruct-v0.2}}. We chose this model as, despite its modest size, it performs well on various natural language tasks and even outperforms its larger counterparts, namely Llama~1 and Llama~2, on some tasks \cite{Jiang_23}. Given this, the model is particularly suitable for our use case where computational resources are limited.
In all LLM-related tasks, we constrained the LLMs' outputs to follow JavaScript Object Notation (JSON) grammar by using the decoding framework of \citet{Geng_23}, who demonstrate improved performance from enforcing more structured outputs.


\paragraph{Aspect Extraction}
To generate a list of aspects from the review texts, we adopted a fine-tuning approach. In Appendix~\ref{appx:aspect_comparison} we report a comparison with prompt-based approaches, which underperformed. For the fine-tuning, we used the SemEval-2014 Task 4 dataset \cite{Pontiki_14}, specifically the manually annotated ground-truth test set from the two provided domains, laptops and restaurants, consisting of 1,600 samples, split evenly. We provide training details in Appendix~\ref{appx:aspect_training}.

During fine-tuning, we noticed that the LLMs had a tendency to generate aspects which were not present in the reviews. These hallucinations could have introduced irrelevant aspects into the ontology. It also tended to misidentify descriptive adjectives as aspects. Therefore, we filtered the extracted aspects by tokenising the review text and then identifying the grammatical roles of each word through part-of-speech (POS) tagging using the Natural Language Toolkit (NLTK) library\footnote{\href{https://www.nltk.org}{https://www.nltk.org}}. Only nouns identified by the POS tagging which were present in the review text were included in the final set of aspects. \DZ{While this filtering 
resulted in fewer hallucinations in our experiments, 
\AR{as LLMs improve with time,} 
\AR{this trade-off (due to the resulting loss of information)} would ideally no longer be necessary.}
Finally, we selected the top 50 most common aspects from the filtered list, then we determined (by visual inspection) that the most important aspects were contained within this number. Doing so also helped to eliminate overly specific aspects, as these tended to occur less frequently in the general reviews and were naturally filtered out.

\paragraph{Synset Extraction}
For grouping the aspects into synsets, we used Equidistant Nodes Clustering (ENC) on the cosine similarity between the word embeddings of the aspects, which achieves state-of-the-art performance \citep{chernoskutov2018equidistant, Oksanen_21}. We also explored K-Means clustering, which we found produced overly large clusters. We report a comparison of these two methods in Appendix~\ref{appx:synset_comparison}.

To produce the word embeddings, we fine-tuned the FastText model (details in Appendix~\ref{appx:synset_training}) due to its resilience to noisy data \cite{Bojanowski_17}. This is especially crucial in the context of product reviews, which tend to have a higher proportion of misspelled words and abbreviations. 
Furthermore, unlike the widely used Word2Vec, which may fail to produce vectors for multi-word phrases with lower occurrence, FastText can construct meaningful representations of these phrases by combining multiple subword units.

\paragraph{Concept Extraction}

From the extracted groups of synsets, we selected the most commonly occurring term of every group as the term to represent the group. These representative aspect terms then formed the set of potential concept candidates. We then prompted the LLM to determine whether each term should be included in the meronymic ontology, and only the synset groups whose representative aspect terms receive a positive response were included in the ontology. The prompt  for this task is \AR{detailed} in Appendix \ref{appx:concept_prompt}. 

\paragraph{Relation Extraction}

The shortlisted synset groups obtained from the previous task form the nodes of the final ontology. To construct the ontology, we extracted the relations between these groups. We did so by isolating sentences containing exactly two aspects from different synset groups. These sentences, along with the two aspects, were then used to prompt the LLM to determine whether a part-whole relationship existed between them (\AR{see} Appendix~\ref{appx:relation_prompt} for \AR{the} prompt). This effectively reduces the task to a multiple-choice problem with answers: 1) the first aspect 
is a part of the second aspect
; 2) the second aspect 
is a part of the first aspect
; and 3) there is no such relationship between the two aspects. \DZ{
\AR{We believe this presents a reasonable starting point for meronomies and leave to future work an investigation of }
more complex relationships between multiple aspects
.} 

We used distillation 
to fine-tune the Mistral model for this task \DZ{on 1000 synthetic samples outputted by Gemini (which was prompted as above) and using the same training settings as \AR{those} in Appendix~\ref{appx:aspect_training}}. We generated these samples from five product categories (backpack, cardigan, camera, guitar and laptop) that are different from those we tested the model on to ensure it does not simply memorise the relationships. 

We have also experimented with using the full reviews and using Mistral without distillation. We report the comparison in Appendix~\ref{appx:relation_comparison}.


\section{Experimental Evaluation}
\label{sec:evaluation}

Since there are no standard benchmarks in meronymic ontology extraction, we decided to evaluate our method in terms of the relevance of the two components it extracts, i.e. the (i) terms and (ii) relations of the final ontology.

We compared against the baseline method proposed by \citet{Oksanen_21}. To ensure a fair comparison, we generated two versions of our ontology: an unedited version (full) and a shortened version (short), which has an equal number of terms as the baseline. This was done by keeping the top-K most commonly occurring terms in our ontology, where K is set to the number of terms in the baseline ontology. All experiments were run on an NVIDIA GeForce RTX 4090 GPU. \DZ{We report \AR{the} mean results over three runs.}

\paragraph{LLMs for Evaluation}
Since user studies are costly for evaluation, we left them to future work and instead turned to LLM-as-a-judge\AR{, as is increasingly being deployed for simple, domain-specific tasks \cite{Huang_25,Wang_25}}. We used Gemini 1.5 Flash\footnote{\url{https://ai.google.dev/gemini-api/docs/models\#gemini-1.5-flash}} \AR{as the LLM judge, to which we}
provided a detailed scoring guide 
that included descriptions of what each score means and examples for each
. The LLM judge then used this guide to output a score from 1 to 5. In addition, the LLM judge was prompted to provide explanations for its scoring as part of the chain-of-thought process. 

To evaluate the terms in the ontology, we used the following criteria (prompt in Appendix~\ref{appx:aspect_eval_prompt}):
    
    \textbf{1. Relevance:} Does the term accurately represent a part or component of the specified product?
    
    \textbf{2. Specificity:} Is the term specific enough to be meaningful within the context of the product, without being overly broad or too narrow?
    
    \textbf{3. Clarity:} Does the term clearly convey the intended part or component, avoiding ambiguity?
    
    \textbf{4. Product Fit:} Is the term logically and contextually appropriate for the given product?

To evaluate the extracted relations, we used the following criteria (prompt in Appendix~\ref{appx:relation_eval_prompt}):
    
    \textbf{1. Logical Hierarchy:} Does the child node represent a logical part, property, or characteristic of the parent node?
    
    \textbf{2. Contextual Fit:} Is the relation reasonable within the context of product categories commonly found on Amazon?
    
    \textbf{3. Clarity and Specificity:} Does the relation avoid ambiguity and clearly define the part-whole or attribute-characteristic relationship? 








\paragraph{Overall Ontology Evaluation}

\begin{table}[t]
    \centering
    \resizebox{\columnwidth}{!}{
    \begin{tabular}{lccc}
        \toprule
        Product & Ours (Full) & Ours (Short) & BERT \\
        \midrule
        Video Game & 4.00 & \textbf{4.18} & 3.92 \\
        Television & \textbf{4.06} & 4.05 & 3.95 \\
        Necklace & 4.50 & \textbf{4.57} & 3.86 \\
        Watch & 4.13 & \textbf{4.37} & 4.10 \\
        Stand Mixer & 4.36 & \textbf{4.40} & 3.31 \\
        \bottomrule
    \end{tabular}
    }
    \caption{Mean average scores of terms extracted from the three different ontologies, as evaluated using an LLM judge, across five products.}
    \label{tb:term_results}
\end{table}

\begin{table}[t]
    \centering
    \resizebox{\columnwidth}{!}{
    \begin{tabular}{lccc}
        \toprule
        Product & Ours (Full) & Ours (Short) & BERT \\
        \midrule
        Video Game & \textbf{3.89} & 3.82 & 3.43 \\
        Television & 3.99 & \textbf{4.56} & 3.21 \\
        Necklace & 3.65 & \textbf{3.79} & 3.29 \\
        Watch & 3.75 & \textbf{4.06} & 2.68 \\
        Stand Mixer & 3.30 & \textbf{3.40} & 2.47 \\
        \bottomrule
    \end{tabular}
    }
    \caption{Mean average scores of relations extracted from the three different ontologies, as evaluated using an LLM judge, across five products.}
    \label{tb:relation_results}
\end{table}

\begin{table}[t]
    \centering
    \resizebox{0.8\columnwidth}{!}{
    \begin{tabular}{lc}
        \toprule
        Stage & Avg Time (min) \\
        \midrule
        Aspect Extraction & 32.05 \\
        Synset Extraction & 0.78 \\
        Concept Extraction & 1.52 \\
        Ontology Extraction & 4.53 \\
        \hline
        Full Pipeline & 38.89 \\
        \bottomrule
    \end{tabular}
    }
    \caption{Average time taken, in minutes, across the five products for each stage in the ontology extraction pipeline for our method.}
    \label{tb:llm_time}
\end{table}

\begin{table}[h!]
    \centering
    \resizebox{0.8\columnwidth}{!}{
    \begin{tabular}{lc}
        \toprule
        Stage & Avg Time (min) \\
        \midrule
        Entity Extraction & 1.66 \\
        Aspect Extraction & 2.79 \\
        Synonym Extraction & 0.82 \\
        Ontology Extraction & 1.36 \\
        \hline
        Full Pipeline & 6.62 \\
        \bottomrule
    \end{tabular}
    }
    \caption{Average time taken, in minutes, across the five products for each stage in the ontology extraction pipeline for the BERT-based method.}
    \label{tb:bert_time}
\end{table}

Table~\ref{tb:term_results} shows that the terms in our shortened ontology generally were judged the best, followed by our full ontology, and finally the BERT ontology. Table~\ref{tb:relation_results} shows similar results for 
relation extraction.


\paragraph{Extraction Time Evaluation}

We first evaluated the performance of the two extraction methods based on the total processing time for the complete extraction of the ontology. We compared the breakdown of the time taken for each stage in the ontology extraction pipeline for our method (Table~\ref{tb:llm_time}) and the BERT-based method (Table~\ref{tb:bert_time}).



\section{Conclusions}
\label{sec:conclusions}
We have proposed an LLM-based approach to extracting meronymic ontologies, demonstrating that our approach significantly improves the relevance of the extracted ontology over a BERT baseline but at the expense of higher computational costs. 

In future work, we plan on investigating if a standard benchmark could be proposed for this task. Moreover, the evaluation of the ontologies could be verified with user studies. 
\AR{Further, we would like to determine how generalisable our approach is to other ontologies, most obviously taxonomies.}

\section{Limitations}
\label{sec:limitations}
The evaluations in this paper rely on LLM-as-a-judge\AR{; supplementing this with user studies may provide more evidence for our findings}. \AR{For example, i}n the real-world, well-organised product ontologies may ultimately be used to drive user engagement. A comprehensive user study could therefore extend to the impact of product ontologies on downstream metrics. \DZ{We note, however, that user studies may suffer from a different set of biases \citep{chen2024humans}.} \DZ{As with most LLM-based approaches, hallucinations are a challenge in \AR{our} work. While a fundamental solution to this problem is outside the scope of this \AR{paper}, we partially mitigate against this by fine-tuning our models on real product ontologies. Future work could also investigate the use of ensembles of LLMs as a way of mitigating hallucinations in any single model.} Product ontologies currently also lack a standard benchmark, with which future progress could be better evaluated. Finally, the performance-improving benefits of LLM-based approaches need to be weighed against potentially higher costs, such as in environmental or monetary terms. 

\section*{Acknowledgments}

Rago was partially funded by The Alan Turing Institute on the UK-Italy Trustworthy AI Visiting Researcher Programme (Award Reference: VRP006).
The authors thank Esmanda Wong for her contributions to this work as part of her Master's thesis. 

\bibliography{custom}

\appendix

\newpage
\quad
\newpage

\section{Aspect Extraction}

\subsection{Training Details}
\label{appx:aspect_training}
We fine-tune the LLM on the 1,600 annotated samples from the SemEval-2014 Task 4 dataset that relate to laptops and restaurants. We set aside 10\% of the data for validation. We use the Adam optimizer \citep{kingma2014adam} with a learning rate of $10^{-4}$ and a cosine scheduler with a warm-up ratio of 0.1. We use LoRA \citep{hu2022lora} applied to linear projection layers with $r=4$ and $\alpha=16$. We train the model for 3 epochs on effective batch sizes of 16 (using gradient accumulation). The model achieved a final accuracy of 0.9909 on the validation set. 

\subsection{Comparison with Prompt-based Approaches}
\label{appx:aspect_comparison}
Besides the fine-tuning approach used in the main paper, we experimented with purely prompt-based approaches (details of the prompts are in Appendix~\ref{appx:aspect_prompt_1} and \ref{appx:aspect_prompt_2}). We used LLM-as-a-judge to evaluate the terms as in the main paper.

Table~\ref{tab:aspect_eval} reports the mean scores given by the LLM judge across three runs, showing that the fine-tuning approach performed best.

\begin{table}[ht]
\centering
\begin{tabular}{lc}
\toprule
Method & Average \\
\midrule
Method A1 & 1.960 ± 0.006 \\
Method A2 & 2.259 ± 0.002 \\
Method A3 & 2.662 ± 0.006 \\
\bottomrule
\end{tabular}
\caption{Evaluation of unique aspects generated by different methods on 1000 necklace reviews.}
\label{tab:aspect_eval}
\end{table}

\subsubsection{Method A1 Prompt: Generating Aspects Only} 
\label{appx:aspect_prompt_1}

\texttt{
You are provided with customer reviews of various products from Amazon. Your task is to identify and extract specific aspects of the product mentioned in each review. Each aspect refers to a particular feature, attribute, or component of the product. Identify aspects using the exact words used in the review, do not make your own aspects.
}

\texttt{
[Start of Examples] \\
Review: In the shop, these MacBooks are encased in a soft rubber enclosure - so you will never know about the razor edge until you buy it, get it home, break the seal and use it (very clever con). \\
Output: \{``aspects": [``rubber enclosure", ``edge"]\} \\
Review: I investigated netbooks and saw the Toshiba NB305-N410BL. \\
Output: \{``aspects": []\} \\
Review: Great laptop that offers many great features! \\
Output: \{``aspects": [``features"]\}
\newline
[End of Examples]
}

\texttt{
[Start of Review] \\
\{INSERT REVIEW HERE\}
\newline
[End of Review]
}

\subsubsection{Method A2 Prompt: Generating Aspects with Sentiment Polarity} \label{appx:aspect_prompt_2}

\texttt{
You are provided with customer reviews of various products from Amazon. Your task is to identify and extract specific aspects of the product mentioned in each review and label the sentiment associated with each aspect. Each aspect refers to a particular feature, attribute, or component of the product. The sentiment can be classified as positive, negative, or neutral. Identify aspects using the exact words used in the review, do not make your own aspects.
}

\texttt{
[Start of Examples] \\
Review: In the shop, these MacBooks are encased in a soft rubber enclosure - so you will never know about the razor edge until you buy it, get it home, break the seal and use it (very clever con). \\
Output: \{``aspects": [\{``aspect": ``rubber enclosure", ``polarity": ``positive"\}, \{``aspect": ``edge", ``polarity": ``negative"\}]\} \\
Review: I investigated netbooks and saw the Toshiba NB305-N410BL. \\
Output: \{``aspects": []\} \\
Review: Great laptop that offers many great features! \\
Output: \{``aspects": [\{``aspect": ``features", ``polarity": ``positive"\}]\}
\newline
[End of Examples]
}

\texttt{
[Start of Review] \\
\{INSERT REVIEW HERE\}
\newline
[End of Review]
}

\section{Synset Extraction}
\subsection{Word Embedding Model}
\label{appx:synset_training}
We trained the model on the full dataset of 100,000 review texts (26,464 for stand mixers) using the skipgram algorithm. The model was trained with a window size of 5 and a vector size of 100 for the word embeddings. We use the same fine-tuning settings as in Appendix~\ref{appx:aspect_training}.

\subsection{Comparison with Alternative Clustering Method}
\label{appx:synset_comparison}
Besides ENC, we also tested K-means clustering of the word embeddings. To tune the choice of $k$, we searched $k \in [10,40]$ to find the highest silhouette score amongst the resulting clustering outcomes. It is evident from Table~\ref{tab:synset_comparison} that the ENC algorithm was much stricter in its grouping of terms, resulting in more but sparsely populated clusters, containing terms with high semantic similarity. In contrast, K-means tended to produce overly large clusters that grouped together terms representing distinct concepts, leading to less precise clusters.

\begin{table}[ht]
\centering
\resizebox{\columnwidth}{!}{
\begin{tabular}{lcccc}
\toprule
\textbf{Product} & \multicolumn{2}{c}{\textbf{ENC}} & \multicolumn{2}{c}{\textbf{K-Means}} \\ \cmidrule(lr){2-3} \cmidrule(lr){4-5}
                 & \textbf{\# Synsets} & \textbf{Avg Size} & \textbf{\# Synsets} & \textbf{Avg Size} \\
\midrule
Video Game    & 33 & 1.52 & 20 & 2.50 \\
Television    & 32 & 1.56 & 21 & 2.38 \\
Necklace      & 30 & 1.67 & 28 & 1.79 \\
Watch         & 39 & 1.28 & 16 & 3.13 \\
Stand Mixer   & 36 & 1.39 & 13 & 3.85 \\
\bottomrule
\end{tabular}
}
\caption{Number of synset groups and the average group size.}
\label{tab:synset_comparison}
\end{table}

\section{Concept Extraction Prompt}
\label{appx:concept_prompt}

\texttt{
You are provided with a list of candidate aspect terms related to a specific product in an e-commerce context. The goal is to determine whether each term should be included as part of a product aspect ontology. A product aspect ontology consists of a root entity, which is the product itself, and various aspects that represent features/sub-features or components/sub-components of the product.
}

\texttt{
For each candidate term, evaluate its relevance and appropriateness for inclusion in the ontology. Consider the following guidelines: \\
Relevance: The term must directly relate to a specific feature, sub-feature, component, or sub-component of the product. \\
Specificity: The term should not be overly broad or overly narrow. It must clearly identify a distinct aspect of the product, but able to generalise across multiple products. \\
Hierarchy: Consider whether the term represents a primary feature or a more granular sub-feature/component.
}

\texttt{
For each candidate aspect term, respond with ``yes" if the term should be included in the product aspect ontology and ``no" if it should not be included. Additionally, provide a brief explanation for your decision.
}

\texttt{
[Start of Examples] \\
Product: Smartphone \\
Candidate Aspect: battery \\
Output: {``answer": ``yes"} (Explanation: Battery is a core component of a smartphone, making it a relevant and specific aspect.) \\
Product: Smartphone \\
Candidate Aspect: fast \\
Output: {``answer": ``no"} (Explanation: Fast describes the smartphone's performance, but it is not a component or feature of the smartphone.) \\
Product: Laptop \\
Candidate Aspect: laptop bag \\
Output: {``answer": ``no"} (Explanation: Although related, a laptop bag is an accessory, not a component or feature of the laptop itself.) \\
Product: Laptop \\
Candidate Aspect: apple \\
Output: {``answer": ``no"} (Explanation: Apple refers to a brand of a laptop, but is too specific and does not generalise well across laptop products.) \\
Product: Earrings \\
Candidate Aspect: gift \\
Output: {``answer": ``no"} (Explanation: Although earrings can be a gift, it is not a component or feature of the earring itself.)
\newline
[End of Examples]
}

\texttt{
Product: \{INSERT PRODUCT HERE\} \\
Candidate Aspect: \{INSERT ASPECT HERE\}
}

\section{Relation Extraction}

\subsection{Comparison}
\label{appx:relation_comparison}
Instead of fine-tuning Mistral, we experiment with using the base model and providing it with the same review excerpt as in the main paper and the full review. We evaluated the ontologies generated by the three different methods using LLM-as-a-judge as in the main paper. 

Table~\ref{tab:extract_eval} reports the LLM judge scores across three runs. Fine-tuning (``Excerpt FT'') yields the best performance, a significant improvement from just using the base model (``Excerpt''), suggesting that fine-tuning was effective. Using the base model but providing it with the full review (``Full'') achieves a competitive performance without the need for fine-tuning, presenting a possible approach in a low-resource context. Table~\ref{tab:time_taken}, however, shows that using the full review significantly increases the run time.

\begin{table}[ht]
\centering
\resizebox{\columnwidth}{!}{
\begin{tabular}{lccc}
\toprule
\textbf{Object Type} & \textbf{Full} & \textbf{Excerpt} & \textbf{Excerpt FT} \\
\midrule
Video Game   & 3.811 $\pm$ 0.057 & 3.727 $\pm$ 0.045 & 3.893 $\pm$ 0.029 \\
Television   & 4.155 $\pm$ 0.061 & 3.726 $\pm$ 0.045 & 3.987 $\pm$ 0.036 \\
Necklace     & 3.397 $\pm$ 0.081 & 3.481 $\pm$ 0.026 & 3.646 $\pm$ 0.029 \\
Watch        & 3.570 $\pm$ 0.110 & 3.398 $\pm$ 0.030 & 3.747 $\pm$ 0.016 \\
Stand Mixer  & 3.080 $\pm$ 0.065 & 2.493 $\pm$ 0.019 & 3.303 $\pm$ 0.021 \\
\bottomrule
\end{tabular}
}
\caption{Evaluation of relations from the ontologies.}
\label{tab:extract_eval}
\end{table}

\begin{table}[ht]
\centering
\resizebox{\columnwidth}{!}{
\begin{tabular}{lccc}
\toprule
\textbf{Object Type} & \textbf{Full} & \textbf{Excerpt} & \textbf{Excerpt FT} \\
\midrule
Video Game   & 20.70 & 14.58 & 4.86 \\
Television   & 28.59 & 21.93 & 7.63 \\
Necklace     & 9.46  & 6.28  & 1.97 \\
Watch        & 15.77 & 11.18 & 3.77 \\
Stand Mixer  & 19.94 & 14.20 & 4.44 \\
\bottomrule
\end{tabular}
}
\caption{Time taken for relation extraction (minutes).}
\label{tab:time_taken}
\end{table}

\subsection{Relation Extraction Prompt} 
\label{appx:relation_prompt}

\texttt{
You are provided with a sentence and two aspects extracted from the sentence. Your task is to determine if there is a meronym (part-whole) relationship between these two aspects. A meronym relationship exists when one aspect is a part of another aspect. Use common sense and the sentence as context for the identified relationship, if any.
}

\texttt{
[Start of Examples] \\
Sentence: only a couple gripes cause im picky.. the sunburst color of the finish was a little too dark. \\
Aspect1: finish \\
Aspect2: color \\
Output: \{``meronym": [\{``part": ``color", ``whole": ``finish"\}]\} \\
Sentence: nice to use on vacation when shopping but fought the straps had to put knots in them to stay on my back.not water proof \\
Aspect1: water proof \\
Aspect2: straps \\
Output: \{``meronym": []\} \\
Sentence: great laptop, except for the worst keyboard ever almost everything about this laptop is great. \\
Aspect1: keyboard \\
Aspect2: laptop \\
Output: \{``meronym": [\{``part": ``keyboard", ``whole": ``laptop"\}]\} \\
Sentence: i am also attaching an image taken of a tree in the sunlight so you can see the dynamic range and how the camera handles sun flares.all images are using default camera's settings except i switched to ``fine" compression, the default is ``normal", and no images were post processed. \\
Aspect1: camera \\
Aspect2: settings \\
Output: \{``meronym": [\{``part": ``settings", ``whole": ``camera"\}]\} \\
Sentence: good buy really happy with  the style and color. \\
Aspect1: color \\
Aspect2: style \\
Output: \{``meronym": []\}
\newline
[End of Examples]
}

\texttt{
Sentence: \{INSERT SENTENCE HERE\} \\
Aspect1: \{INSERT ASPECT1 HERE\} \\
Aspect2: \{INSERT ASPECT2 HERE\}
}

\section{Evaluation Prompts}
\label{appx:eval_prompt}

\subsection{Evaluation of Aspect Extraction} \label{appx:aspect_eval_prompt}

\texttt{
You are an AI judge tasked with evaluating the correctness of terms within an Amazon product ontology.
Specifically, you will assess whether a given term is appropriate and correctly categorized as a part, component (meronym) or attribute of a specified product. The terms should be general enough to represent common parts, components or attributes across different listings of the product. The ontology follows a meronymic structure, where terms should represent parts, components or attributes that logically fit within their associated products. Strictly follow this format to output the scores, followed by the explanation: Score: [[1-5]], e.g., Score: [[1]].
}

\texttt{
Evaluation Criteria: \\
Relevance: Does the term accurately represent a part, component or attribute of the specified product?
Specificity: Is the term general enough to be meaningful within the context of the product, avoiding overly specific terms? \\
Clarity: Does the term clearly convey the intended part, component or attribute, avoiding ambiguity? \\
Product Fit: Is the term logically and contextually appropriate for the given product?
}

\texttt{
Score 1: Completely incorrect, irrelevant, or overly specific term for the product. (e.g., Product: Smartphones, Term: Laptop) \\
Score 2: Poorly fitting term with minimal relevance or appropriateness, or overly specific for the product. (e.g., Product: Smartphones, Term: diamond bling phone cover) \\
Score 3: Fairly appropriate term with some relevance, but it may lack specificity, clarity, or perfect fit within the product. (e.g., Product: Smartphones, Term: Box) \\
Score 4: Good term with relevance and a logical fit, but it may have slight ambiguities. (e.g., Product: Smartphones, Term: Features) \\
Score 5: Excellent term that is highly relevant, specific enough, clear, general, and a perfect fit for the product. (e.g., Product: Smartphones, Term: Screen size) 
}

\texttt{
Term to evaluate: \\
Product: \{INSERT PRODUCT HERE\} \\
Term: \{INSERT TERM HERE\}
}

\subsection{Evaluation of Relation Extraction} \label{appx:relation_eval_prompt}

\texttt{
You are an AI judge evaluating the correctness of meronym (part-whole) and attribute (property-characteristic) relations within an Amazon product ontology. Your task is to score the given child-parent node relations based on how well the child node represents a part, property, or characteristic of the specified parent node. For each relation, you will analyze whether the child node logically and hierarchically fits as a part or attribute of the parent node in the context of the product category \{INSERT CATEGORY HERE\}. Strictly follow this format to output the scores, followed by the explanation: Score: [[1-5]], e.g., Score: [[1]].
}

\texttt{
Evaluation Criteria: \\
Logical Hierarchy: Does the child node represent a logical part, property, or characteristic of the parent node? \\
Contextual Fit: Is the relation reasonable within the context of product categories commonly found on Amazon? Consider attributes relevant to listings, but allow flexibility for less common, yet valid, relationships. \\
Clarity and Specificity: Does the relation avoid ambiguity and clearly define the part-whole or attribute-characteristic relationship? Acknowledge general, but correct, relations even if they lack specific detail.
}

\texttt{
Score 1: Completely incorrect relation with no logical or contextual fit. (e.g., Child Node: apple, Parent Node: car) \\
Score 2: Poor relation with minimal logical or contextual fit. (e.g., Child Node: van, Parent Node: bike helmet) \\
Score 3: Fair relation with some logical fit but lacks strong contextual relevance or clarity. (e.g., Child Node: book, Parent Node: school) \\
Score 4: Good relation with a logical and contextual fit but may have slight ambiguities. (e.g., Child Node: features, Parent Node: vehicle) \\
Score 5: Excellent relation with a clear, logical, and contextual fit, with no ambiguities. (e.g., Child Node: chapter, Parent Node: book) \\
}

\texttt{
Relation to evaluate: \\
Child Node: \{INSERT CHILD HERE\} \\
Parent Node: \{INSERT PARENT HERE\}
}

\end{document}

%% file: macros.tex
\newcommand{\AR}[1]{\textcolor{black}{#1}}

\newcommand{\delete}[1]{}
\newcommand{\DZ}[1]{\textcolor{black}{#1}}

%% file: packages.tex
\usepackage{xcolor}
\usepackage{hyperref}
\usepackage{booktabs}
\usepackage{todonotes}